\documentclass{article}

\usepackage{microtype}
\usepackage{graphicx}
\usepackage{booktabs} 

\usepackage{hyperref}

\usepackage{pifont}
\newcommand{\cmark}{\ding{51}}

\usepackage[preprint]{icml2023}
\usepackage[inline]{enumitem}
\usepackage{caption}
\usepackage{subcaption}
\usepackage{amsmath}
\usepackage{amssymb}
\usepackage{mathtools}
\usepackage{amsthm}

\newcommand{\E}{\mathbb{E}}

\newcommand{\Alg}{\mathcal{A}}

\newcommand{\bc}{\mathbf{c}}

\newcommand{\bx}{\mathbf{x}}

\newcommand{\bz}{\mathbf{z}}

\newcommand{\diffusionmodel}{{\boldsymbol{\epsilon}_{\theta}}}

\newcommand{\distillmodel}{\diffusionmodel^{distill}}
\newcommand{\realmodel}{\diffusionmodel^{real}}
\newcommand{\Dreal}{D_{real}}
\newcommand{\Dsyn}{D_{synth}}
\newcommand{\Dfilt}{D_{filtered}}
\newcommand{\Ddistill}{D_{distill}}
\newcommand{\freid}{f_{\theta}^{re-id}}
\newcommand{\fretrieval}{f_{\theta}^{retrieval}}
\newcommand{\fimtext}{f_{\theta}^{im2tex}}
\newcommand{\fclass}{f_{\theta}^{class}}
\newcommand{\salign}{s_{align}}
\newcommand{\sreid}{s_{re-id}}
\newcommand{\Rreid}{R_{re-id}}

\usepackage[capitalize,noabbrev]{cleveref}

\theoremstyle{plain}
\newtheorem{theorem}{Theorem}[section]

\theoremstyle{definition}
\newtheorem{definition}[theorem]{Definition}

\theoremstyle{remark}

\usepackage[textsize=tiny]{todonotes}

\icmltitlerunning{Privacy Distillation:Reducing Re-identification Risk of Multimodal Diffusion Models}

\begin{document}

\twocolumn[
\icmltitle{Privacy Distillation: \\ Reducing Re-identification Risk of Multimodal Diffusion Models}



\icmlsetsymbol{equal}{*}

\begin{icmlauthorlist}
\icmlauthor{Virginia Fernandez}{equal,kcl}
\icmlauthor{Pedro Sanchez}{equal,ue}
\icmlauthor{Walter Hugo Lopez Pinaya}{equal,kcl}
\icmlauthor{Grzegorz Jacenków}{ue} 
\icmlauthor{Sotirios Tsaftaris}{ue}
\icmlauthor{Jorge Cardoso}{kcl}
\end{icmlauthorlist}

\icmlaffiliation{kcl}{King's College London, Strand,
London, United Kingdom}
\icmlaffiliation{ue}{The University of Edinburgh, Old College, South Bridge, Edinburgh, United Kingdom}

\icmlcorrespondingauthor{Virginia Fernandez}{virginia.fernandez@kcl.ac.uk}

\icmlkeywords{Machine Learning, ICML}

\vskip 0.3in
]



\printAffiliationsAndNotice{\icmlEqualContribution} 

\begin{abstract}
Knowledge distillation in neural networks refers to compressing a large model or dataset into a smaller version of itself. We introduce \emph{Privacy Distillation}, a framework that allows a text-to-image generative model to teach another model without exposing it to identifiable data. Here, we are interested in the privacy issue faced by a data provider who wishes to share their data via a multimodal generative model. A question that immediately arises is ``\textit{How can a data provider ensure that the generative model is not leaking identifiable information about a patient?}''. Our solution consists of (1) training a first diffusion model on real data (2) generate a synthetic dataset using this model and filter it to exclude images with a re-identifiability risk (3) train a second diffusion model on the filtered synthetic data only.
We showcase that datasets sampled from models trained with Privacy Distillation can effectively reduce re-identification risk whilst maintaining downstream performance.
\end{abstract}

\section{Introduction}
Synthetic data have emerged as a promising solution for sharing sensitive medical image data \cite{jordon2020synthetic}. By utilising generative models, artificial data can be created with statistical characteristics similar to the original training data, thereby overcoming privacy, ethical, and legal issues that data providers face when sharing healthcare data \cite{jordon2020synthetic,Yoon2020,murtaza2023synthetic}. Recent advancements have made text-to-image generative models such as Stable Diffusion \cite{rombach2021highresolution}, DALL-E \cite{ramesh21a_dalle1,ramesh2022hierarchical} and Imagen \cite{imagen2022} achieve sufficient quality to accurately represent the original data in terms of both realism and diversity.  Beyond text conditioning, they also cope well with various other modalities such as segmentation masks, contours and other spatial information \cite{Zhang2023}. Sharing trained generative models, in particular, can be useful\footnote{As seen in Stable Diffusion's successful public release followed by over 6 million downloads (by March 2023) of its weights by the community \url{https://stability.ai/blog/stable-diffusion-public-release}} for fine-tuning in smaller datasets \cite{ruiz2022dreambooth,Chambon2022RoentGen}, anomaly detection \cite{Pinaya2022anomaly,Sanchez2022WhatIH}, or even leveraging synthetic data for downstream tasks such as segmentation \cite{Fernandez2022sashimi} or classification \cite{li2023synthetic,azizi2023synthetic}.

However, generating high quality synthetic data which are useful for downstream tasks is not enough to diminish the privacy risks on medical data. There is a growing concern on whether deep generative models preserve privacy \cite{GAN_leakage_hitaj,Chen2021}. Deep generative models are prone to leak information about their training datasets \cite{Jegorova2022tpami}. 

A major risk in healthcare is the potential for \textbf{patient re-identification} from the training dataset \cite{Yoon2020}, especially when sharing models derived from private or protected datasets. Re-identification in the context of generative modelling refers to a synthetic image which contains identifiable information about a patient in the training set. Identifiable information is any information that can be used to identify an individual\footnote{The specific definition of ``identifiable information'' can differ between laws and countries.}. The notion of what constitutes a person's identity might be ambiguous in the context of synthetic, anonymised data. However, it has been shown that it's possible for deep learning models to determine whether two images belong to the same patient \cite{Packhauser2022}, even when these images were acquired at different times and when the patient's clinical condition has changed. A potential attacker, with incomplete information about a patient, who manages to trace a synthetic image back to this patient, could learn sensitive clinical information from the image or from the selection criteria of the dataset used to train the generative model \cite{Zhang2022}.



In this work, we propose a distillation procedure where two diffusion models are trained sequentially. The first model is trained on real data and used to generate a synthetic dataset. Subsequently, the synthetic dataset is filtered by a re-identification network to eliminate images that could potentially be used to re-identify real patients. A second model is then trained on the filtered synthetic dataset, thus avoiding the risk of memorisation of the real images and subsequent potential re-identification of patients. The efficacy of the distilled model is evaluated by assessing the performance of a downstream classifier on the synthetic data generated by the distilled model.
Our main \textbf{contributions} are:
\begin{enumerate}
    \item We train a conditional latent diffusion model (LDM) on text-image pairs from a Chest X-ray dataset, following the strategy in RoentGen \cite{Chambon2022RoentGen};
    \item We assess re-identification risk of LDMs trained with different dataset sizes as well as how risk varies when the model is trained from scratch as opposed to fine-tuned;
    \item We propose a distillation procedure which improves privacy and verify that the distilled model has lower re-identification risk, whilst retaining information about the original dataset useful for classifiers on its generated data.
\end{enumerate}

\section{Background and Related Works}

\subsection{Diffusion models for medical images synthesis.} 

Diffusion probabilistic models~\cite{Ho2020DenoisingModels,songscoreSDE,Dhariwal2021} (DPMs) learn to reverse a sequential image noising process, thus learning to map a pure noise image into a target data distribution. Diffusion models can, therefore, be used as a generative model. A particular type of DPM that has been successfully \cite{kazerouni2022diffusion} applied to medical imaging \cite{Chambon2022RoentGen,pinaya2022brain} are latent diffusion models \cite{rombach2021highresolution} (LDMs). LDMs allow the generation of high-dimensional high-resolution images by having a diffusion model over the latent space of a variational autoencoder. Generative modelling in a lower dimensional space allows better conditioning on text \cite{Chambon2022RoentGen} and scale particularly well to high resolution and 3D images \cite{pinaya2022brain}. 

In this paper, we follow RoentGen \cite{Chambon2022RoentGen} in fine-tuning a LDM \cite{rombach2021highresolution} pre-trained\footnote{\url{https://huggingface.co/runwayml/stable-diffusion-v1-5}} on a subset of the LAION-5B database \cite{schuhmann2022laionb}. The latent diffusion model is conditioned on a text $\bc$ which is passed through a text encoder $\tau_{\phi}$. $\tau_{\phi}$ is a pretrained CLIP text encoder \cite{radford2021learning}. The image latent space is obtained from a encoder $\bz = E_{\psi}(\bx)$ which is pretrained along with a decoder $D_{\psi}$ using kullback-leibler (KL) divergence, LPIPS perceptual loss and patch discriminator as described in \citet{rombach2021highresolution}. The latent diffusion model can be implemented with a conditional denoising U-Net $\diffusionmodel(\bz_t, \tau_{\phi}(\bc), t)$ which allows controlling the synthesis process through inputs $\bc$.

Here, we only train the parameters $\theta$ from the diffusion model, leaving the weights $\psi$ and $\phi$ from the autoencoder and text encoder respectively as pre-trained \cite{Chambon2022RoentGen}. Whenever we mention samples from an unconditional model, we refer to images generated with prompts from empty strings.
The training procedure is done by learning a $\theta^*$ such that
\begin{equation}
\label{eq:trainingDDPM}
\theta^* = \underset{\theta}{\arg \min } ~ \E_{\bx_0, t, \epsilon} \left[ \left\| \diffusionmodel(\bx_t, \bc, t)- \epsilon \right\|_{2}^{2}\right],
\end{equation}
where $\bz_t = \sqrt{\alpha_t}\bz_0 + \sqrt{1 - \alpha_t} \epsilon$, with $\bz_0  = E_{\psi}(\bx)$, $t \sim \mathcal{U}\left(0, T\right)$ and $\epsilon \sim \mathcal{N}\left(0, \mathrm{I}\right)$ is the noise. We generate images using classifier-free guidance \cite{ho2021classifier} with the PNDM sampling strategy \cite{liu2022pseudo}.

\subsection{Sample-level Metrics for Synthetic Data}
As detailed in Section \ref{sec:privacy_distillation}, our method involves a filtering procedure. Evaluating the quality of synthetic examples is challenging task and numerous methods such as Inception Score (IS) \cite{9423425}, Frechet Inception Distance (FID) \cite{10.5555/3295222.3295408,ImageNet_FID}, and Precision/Recall (PR)\cite{NEURIPS2019_0234c510}. IS, FID and PR are methods that compute characteristics of the distribution of the synthetic data. If the downstream use case for the synthetic data is defined, one might use downstream performance to evaluate the dataset.

However, in certain scenarios, one is interested in evaluating the qualities of individual synthetic samples such that requirements (such as privacy) over the generated dataset can be enforced post-hoc (after training). Therefore, a \citet{pmlr-v162-alaa22a} explored $\alpha$-precision, $\beta$-recall and authenticity that characterizes the fidelity, diversity and generalisation per sample. \citet{han2023rarity} proposed the ``rarity score'' which measures the uncommonness of generated images using the nearest-neighbor distance in the latent space from other real and synthetic data points. In the multimodal test-to-image setting, a common metric is the CLIP score \cite{ramesh21a_dalle1} which measures the alignment between the conditioning and the generated image.

\subsection{Diffusion Models and Privacy}

DPMs have shown to be particularly susceptible to attacks extracting its training data \cite{Carlini2023ddpm,Somepalli2023Forgery}, exceeding the number of images extracted from other architectures such as generative adversarial networks (GANs) \cite{Carlini2023ddpm}. Few publications have tackled solutions for  privacy preservation in diffusion models. \citet{Carlini2023ddpm} did a thorough analysis of the impact of model hyperparameters, duplicates and training dataset size on the extraction of training samples of two state-of-the-art diffusion models. \citet{Carlini2023ddpm}, however, only measured memorisation via pixel-level similarity (a modified version of $l_{2}$ loss). They retrieve the samples from the training dataset that were closer to a specific synthetic image. 

A popular solution to tackle privacy in deep learning models is the use of differential privacy (DP) \cite{dwork2014algorithmic,abadi2016deep} training. DP in deep learning is performed via differentially private stochastic gradient descent (DP-SGD) \cite{abadi2016deep}. DP-SGD preserves privacy by clipping and noising the parameter gradients during
training. \citet{Dockhorn2022} show how DP-diffusion models generate images of substantially better quality than DP-GAN counterparts and have a much more stable training regime. Nonetheless, the paper showcases that, at present, this approach is limited to models with a small number of parameters, which leaves out its application to large, state-of-the-art models. In addition, despite obtaining outstanding results compared to other DP generative models, the visual quality of these samples is still far from that obtained with non-DP models.

\section{Privacy Distillation}
\label{sec:privacy_distillation}
\subsection{Problem Statement}

Consider a real dataset $\Dreal = \{( \bx_i, \bc_i) \mid \forall i \in (1,2, \dots ,N)\}$ of images $\bx_i$ and text $\bc_i$ belonging to a patient $\mathbf{p}_i$. We are interested in training a generative model $\diffusionmodel$ which is able to synthesise images $\hat{\bx}_i$ such that $\hat{\bx}_i$ does not contain information that can be used to identify a data point $\bx_i, \forall i \in (1,2, \dots ,N)$. 

Following the literature \cite{Carlini2023ddpm,Yoon2020}, we hypothesise that synthetic images can enable re-identification due to model memorisation. 
\begin{definition}[$\ell,\delta$-Memorisation, adapted from \cite{Carlini2023ddpm}]
A $\bx_i$ is considered ($\ell,\delta$)-\textit{memorised} by $\diffusionmodel$ if $\ell(\hat{\bx}_i,\bx_i) \geq \delta$, where $\ell$ is a similarity function, $\delta$ is a threshold, and $\Alg$ is an algorithm which can extract an image $\hat{\bx}_i$ from a generative model $\diffusionmodel$ without access to the original $\bx_i$, $\hat{\bx}_i = \Alg(\diffusionmodel)$. In the case of LDMs, $\Alg$ is a sampling algorithm.
\end{definition}


We assume that $\bc_i$ does not contain identifiable information about $\mathbf{p}_i$, therefore, we only focus on $\hat{\bx}_i$ for identifiable information. This is a reasonable assumption since identifiable information in text, such as demographics, can be easily recognised whereas images can have more subtle details. In the next sections, we will consider two cases where we perform 
\begin{enumerate*}[label=(\roman*)]
    \item conditional sampling $\hat{\bx}_i = \Alg(\diffusionmodel, \bc_i)$;
    \item unconditional sampling $\hat{\bx}_i = \Alg(\diffusionmodel)$.
\end{enumerate*}


\subsection{Distillation Procedure}

\begin{figure*}[t]
\includegraphics[width=0.99\textwidth]{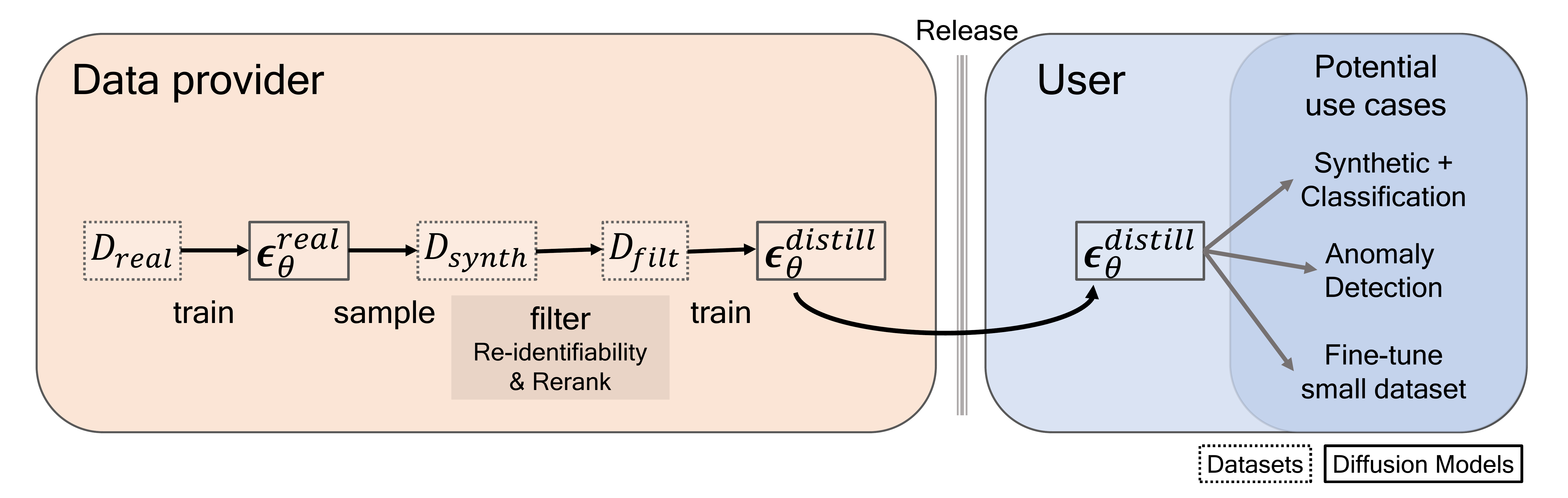}
\caption{Privacy Distillation Pipeline.}
\label{pipeline}
\end{figure*}

As we are interested in safely sharing the weights $\theta$ of generative models in a privacy-preserving manner, a major concern is that synthetic images generated by a model can be used to re-identify a patient from the real training dataset. Therefore, we propose an algorithm for training a diffusion model over filtered synthetic images, minimising the model's exposure to re-identifiable data. 

The procedure for \emph{Privacy Distillation}, as depicted in Figure \ref{pipeline}, consists of the following steps:
\begin{enumerate}
    \item train a diffusion model $\realmodel$ on real data $\Dreal$
    \item generate a synthetic dataset $\Dsyn$
    \item filter $\Dsyn$, ensuring that none of the images are re-identifiable, to obtain $\Dfilt$
    \item train a diffusion model $\distillmodel$ on $\Dfilt$
    \item share $\distillmodel$.
\end{enumerate}

\subsection{Filtering for privacy}



Defining an appropriate $\ell(\hat{\bx},\bx)$ allows controlling which aspects of the original data one wishes to measure for memorisation. Previous work \cite{Carlini2023ddpm} searches near-identical images utilising a Euclidean distance or pixel-by-pixel correspondence. Measuring \emph{identity}, however, can be challenging and specific to certain modalities or organs \cite{KUMAR2017242}, limiting the validity of such approaches. 

\noindent\textbf{Assessing re-identification.~}
Instead of pixel-based \cite{Carlini2023ddpm} or structural-based \cite{KUMAR2017242} similarities, we measure identity with a deep model $\ell = \freid$, introduced by Packhäuser et al. \cite{Packhauser2022}. The model is trained to classify images as belonging to the same patient or not. This model, devised for X-ray images, consists of a siamese neural network with a ResNet-50 backbone. The model takes in two images, fuses the representation of the two branches and outputs a \textit{re-identification score} (after a sigmoid) that we will note as $s_{re-id}$. The model is trained on real images. 

When performing filtering, we compare a real image to a synthetic image. If $s_{re-id} \geq \delta$ for a pair of synthetic and real images $(\hat{\bx}_i,\bx_i)$, we consider that $\hat{\bx}_i$ contains identifiable information about $\bx_i$. For a set of synthetic images, we call \textit{re-identification ratio} $\Rreid$ the number of synthetic samples containing identifiable information about real samples divided by the total number of synthetic samples generated.

We train $\freid$ from scratch on our training set, sampling positive (images from the same patient) and negative pairs, which are randomly sampled. To avoid data imbalance, positive pairs, which were on average ten times less frequent, were oversampled, resulting in an effective dataset size of 472,992. We tested it on a set of even 101,592 non-repeated pairs, achieving a 99.16\% accuracy (AUC 0.9994).


\noindent\textbf{Retrieval.~} 
In a scenario where the text conditioning is not available, we need to search for the most similar real image in the training dataset before computing the re-identification score $s_{re-id}$. Therefore, for data sampled without conditioning, we utilised a retrieval model $\fretrieval$ which was also proposed in \cite{Packhauser2022}. The model works as a feature extractor for computing the nearest neighbours in the embedding space. The model is a siamese neural network with an architecture similar to $\freid$. The $\fretrieval$ excludes the layers from the merging point onwards, to function solely as a feature extractor. $\fretrieval$ is trained with a contrastive loss function.

During filtering, $\fretrieval$ identifies the closest image in terms of identity by computing the Euclidean distance between the embeddings of the query synthetic image and the embedding of every real image in our training set. When evaluating pairs of the test set from the real dataset, our trained model obtained high mean average precision at R (mAP@R) of about 95\% and a high Precision@1 (the precision when evaluating how many times the top-1 images in the retrieved lists are relevant) of 97\%. This way, approach enabled us to analyze and evaluate the unconditioned synthetic data accurately. 

\noindent\textbf{Constrative Reranking.~} 
We also need to ensure that the images in $\Dfilt$ used for training $\distillmodel$ correspond to their conditioning. Therefore, we rerank the synthetic images in $\Dfilt$ based on the image alignment with the conditioning, similar to DALL-E \cite{pmlr-v139-ramesh21a}. 

We leverage CXR-BERT \cite{boecking2022making} for the text encoder which is a chest X-ray (CXR) language model that utilises an improved vocabulary, pretraining procedure and text augmentations tailored to medical text. The model is fine-tuned with a contrastive text-image loss, together with an image encoder \cite{boecking2022making}. An alignment score $\salign = \fimtext(\hat{\bx}_i, \bc_i)$ is computed between an image and a text prompt by passing them through an image and text encoder respectively and taking the cosine similarity between their latent spaces.

\noindent\textbf{Filtering strategy.~} 
We generate $N_c$ synthetic images for each prompt $\bc_i$ in $\Dreal$. We generally choose $N_c = 10$. Therefore, $\Dsyn$ has $N_c * N$ elements. We compute $\salign$ between all generated images and corresponding conditioning using $\fimtext$; and $\sreid$ between the generated images and the real image corresponding to its prompt. For unconditional models, we use $\fretrieval$ to find the strongest candidate in the dataset before computing $\sreid$. We remove all re-identified images $\sreid \geq \delta$ and choose, for each $\bc_i$, the synthetic image with the highest $\salign$.

\section{Experiments}

First, we evaluate how/when identity memorisation happens and the effect of training the model and sampling under different conditions and dataset sizes. Then we showcase that our model trained under Privacy Distillation can be used to train a downstream classification model while reducing re-identification risk.

\subsection{Data}

We use images and radiological reports from the MIMIC-CXR 2.0.0 database \cite{Johnson2019}. As text, we use each report's ``impression'' section, which corresponds to an interpretative summary of the findings for supporting medical decisions. Following RoentGen \cite{Chambon2022RoentGen}, we filter the data used in this study based on the length of impression in tokens, which should not exceed 76 tokens due to the text encoder limit.

Ultimately we obtained a set of 45,453 images belonging to 25,852 patients, each associated with an impression of the original radiological report. We split these into a train set of 23,268 patients (40,960 images) and a test set of 2,132 patients (3,101 images). 10\% of the patients left for testing had half of their images and report pairs moved to the training dataset to allow us to assess re-identification when the patient, but not the query (image, text) pair, is part of the training dataset.

\subsection{Metrics}
\label{downstream}
Beyond using the re-identification score $s_{re-id}$ and the text-to-image similarity $\salign$ for measuring the quality of our synthetic dataset, we also measured image fidelity and downstream performance. We evaluate the fidelity of the synthetic images using the distribution-based metric Fréchet Inception Distance (FID) score \cite{heusel2017gans}. We utilise features extracted by the pre-trained DenseNet-121 from the \textit{torchxrayvision} package \cite{Cohen2022xrv}. 

To assess the quality of synthetic datasets, we train a classifier $\fclass$ of 5 different pathologies (Cardiomegaly, Edema, Consolidation, Atelectasis, Pleural Effusion) based on the model ranked first in the CheXpert Stanford ML leaderboard \cite{Yuan}.\footnote{We used their code, available at \url{https://github.com/Optimization-AI/LibAUC}} We trained a DenseNet-121 on our datasets and tested it in the real hold-out test set. The network is pre-trained for 5 epochs on cross-entropy loss, then trained for another 5 epochs on an AUC loss, as per \cite{Yuan}.

\subsection{Measuring Re-identification Risk of Latent Diffusion Models}

\begin{figure}[h]
\centering
\includegraphics[width=\columnwidth]{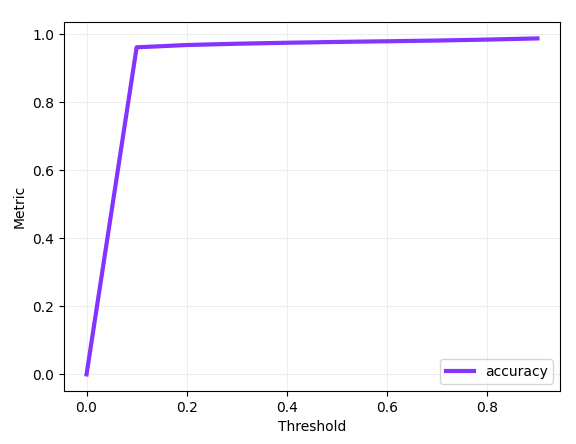}
\caption{Effect of varying threshold $\delta$ on the re-identification score $s_{re-id}$.}
\label{fig:threshold}
\end{figure}

\noindent\textbf{Effect of varying the $\delta$.~}
We explored how $\delta$ threshold for the $s_{re-id}$ score impact the decision if a synthetic data point contains identifiable information of not. We explored the effect of varying $\delta$ and found no relevant difference in the resulting score for thresholds between 0.05 and 0.90, as can be seen by Figure \ref{fig:threshold}. Therefore, we set $\delta = 0.5$ for the rest of the experiments.

\noindent\textbf{Effect of fine-tuning.~}
We explored the differences in terms of $\salign$ between fine-tuning the model pre-trained on LAION-5B or training from scratch, and between sampling using conditioning or not. For the conditioned generation, we sample 100 instances for each of the first 400 prompts of the training dataset, resulting in 40,000 samples. For the unconditional generation, we sample 40,000 images and use the retrieval network to get the closest images in the training dataset. We calculate the re-identification ratio and FID as shown in Table \ref{tab:pretraining_cond}.
The lowest re-identification ratio was achieved for the data sampled from a model trained from scratch using conditioning. Unconditionally-generated datasets have higher re-identification ratios but achieve a better FID score. Nonetheless, their usability is limited, as conditional sampling allows the user to guide the generation process. 
\begin{table}[t]
    \centering
    \caption{Evaluating the influence of pre-training and conditioning on $\realmodel$.}
    \begin{tabular}{ cccc } 
     \toprule
     Pre-trained & Conditional & $\Rreid$ $\downarrow$ & FID $\downarrow$ \\ 
     \midrule
     - & - & 0.057 $\pm$ 0.232 & 54.56 \\ 
     - & \cmark & 0.015 $\pm$ 0.124  & 81.95 \\ 
     \cmark & - & 0.034 $\pm$ 0.181 & 97.91 \\ 
     \cmark & \cmark & 0.022 $\pm$ 0.232 & 79.27 \\ 
     \bottomrule
    \end{tabular}
    \label{tab:pretraining_cond}
\end{table}

\begin{figure*}[h!]
\centering
\includegraphics[width=0.99\textwidth]{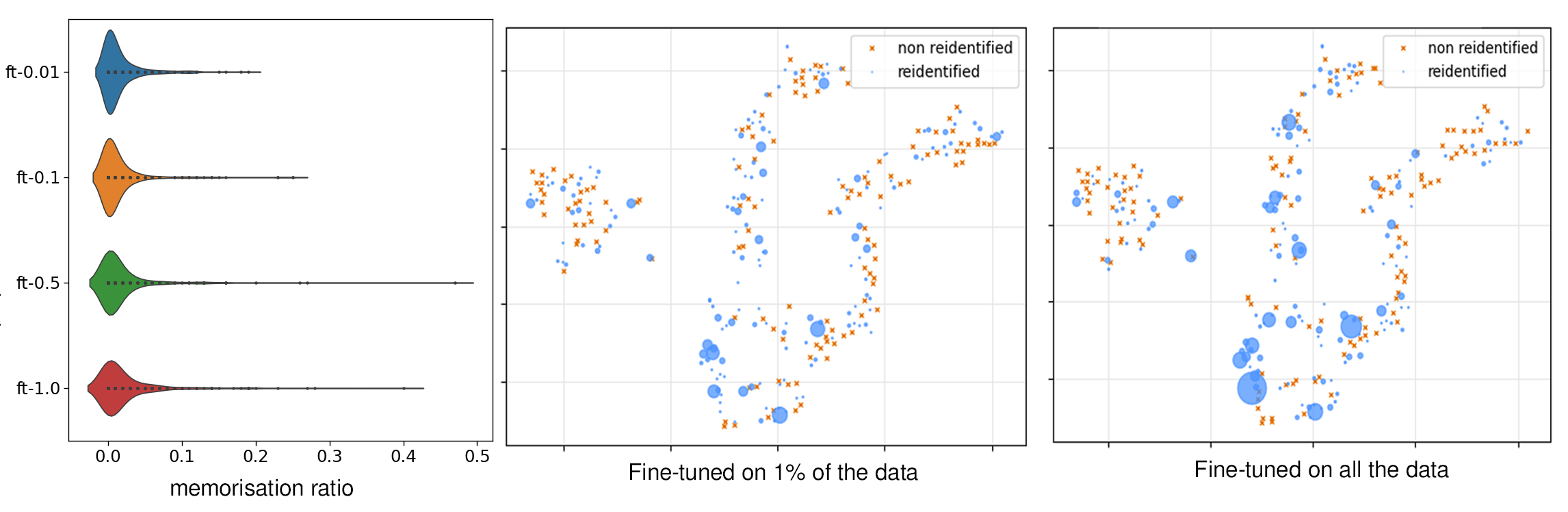}
\caption{Left: violin plots showing the distribution of the average re-identification ratio for the models fine-tuned in different portions of the training dataset; the middle and right plots are TSNE plots of the $\fimtext$ embeddings of the first 400 prompts used to test this experiment, for the model trained on 1\% and 100\% of the data, respectively. Orange dots correspond to prompts for which none of the generated 100 instances was re-identified, whereas blue dots are associated with prompts re-identified to some extent, the size being proportional to the re-identification ratio. }
\label{violin_plots}
\end{figure*}

\noindent\textbf{Effect of training dataset size on memorisation.~}
We trained one model on our full training dataset, and three models on 1\%, 10\% and 50\% of the training dataset, respectively. Then, we calculated the $R_{re-id}$ of a set of samples inferred using 100 instances of 400 training prompts (40,000 images in total). Figure \ref{violin_plots} shows the \textit{re-identification ratio}, defined as the average number of times a sample was re-identified divided by the total of generated samples. 

As opposed to findings in the literature, where bigger training set sizes result in less leakage \cite{Carlini2023ddpm}, the re-identification ratio was lowest for the model trained on only 1\% of the data. The TSNE plots of Figure \ref{violin_plots} suggest that re-identification tends to happen in specific clusters, which aligns with the findings in \cite{Ruolin2022miccai}. We looked at the radiological reports of the top 10 most re-identified prompts, and 90\% of them were associated with similar pathological phenotypes: atelectasis, pleural effusion, and lung opacity. In parallel, we observed that the proportion of images associated with these pathologies is less frequent in the first 1\% of the data than it is in the whole dataset, which suggests that generated data from models trained on the full dataset might be more re-identifiable due to overfitting to subject-specific pathological features. This hypothesis is corroborated by using the Grad-CAM activations of the verification network; we observed that, in patients with pleural effusion or atelectasis, the most salient regions matched the areas with evidence for these pathologies. This, combined with the potential overfitting, likely explains the increase in the average memorisation ratio.

\begin{figure}[h!]
\centering
\subfloat{\includegraphics[width=0.99\columnwidth]{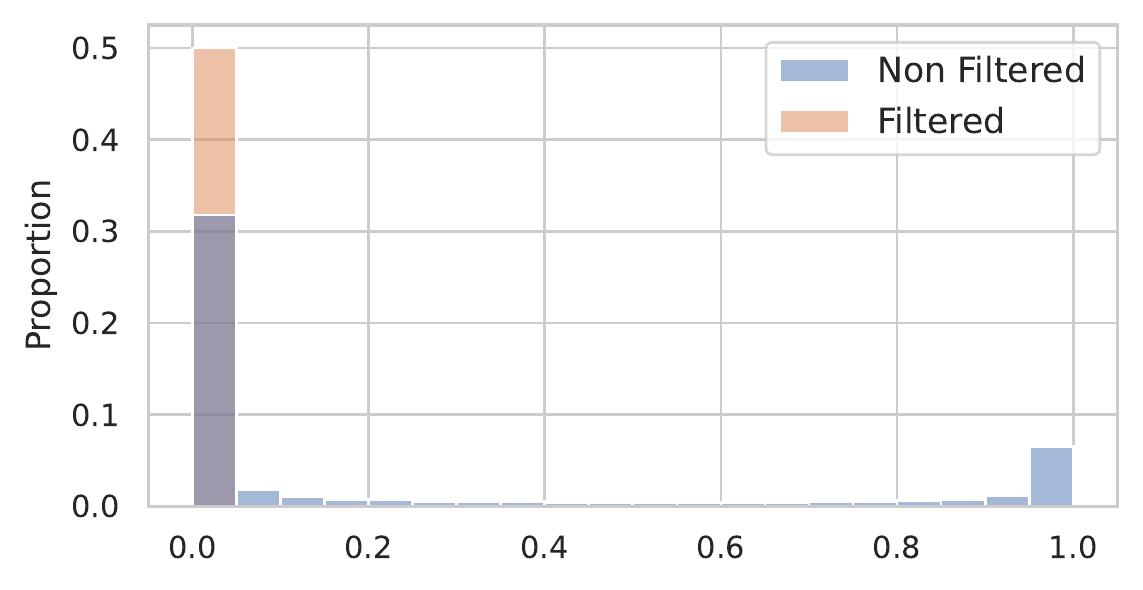}\label{fig:filtered_verification_score}}
\hfill
\subfloat{\includegraphics[width=0.99\columnwidth]{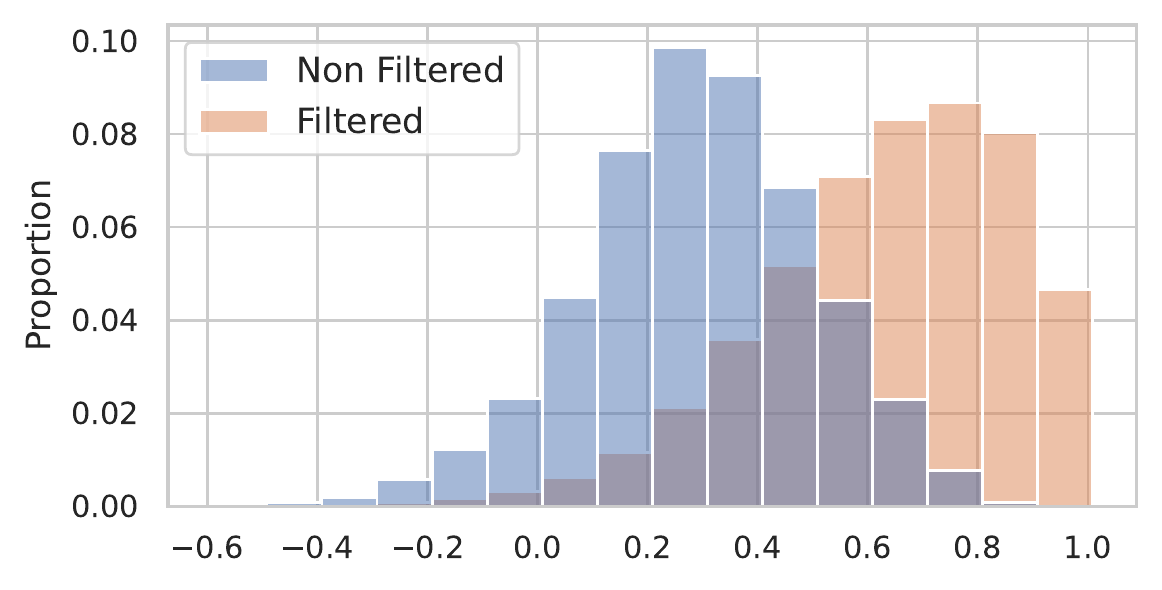}\label{fig:filtered_im2text_similarity}}
\caption{\textit{Top}: $s_{re-id}$ distribution ; \textit{Bottom}: $\salign$ distribution. Comparing distribution of the non-filtered $\Dsyn$ and filtered $\Dfilt$ dataset.}
\label{fig:side}
\end{figure}

\noindent\textbf{Effect of filtering.~}
We compare the impact of our filtering strategy on the synthetic datasets $\Dsyn$ and $\Dfilt$. We sample 10 instances for each of the 40,959 training prompts from the proposed privacy-preserving model. We then filter by $s_{re-id}$, and pick the sample with the better $\salign$ score, resulting in a filtered dataset of 40,959 images (Figure \ref{fig:side}). We found that filtering improves the $\salign$ and reduces the number of memorised (re-identifiable) samples to $0$, given a $\delta$. 

\begin{figure*}[h!]
\centering
\begin{subfigure}[t]{\textwidth}
    \includegraphics[width=\textwidth]{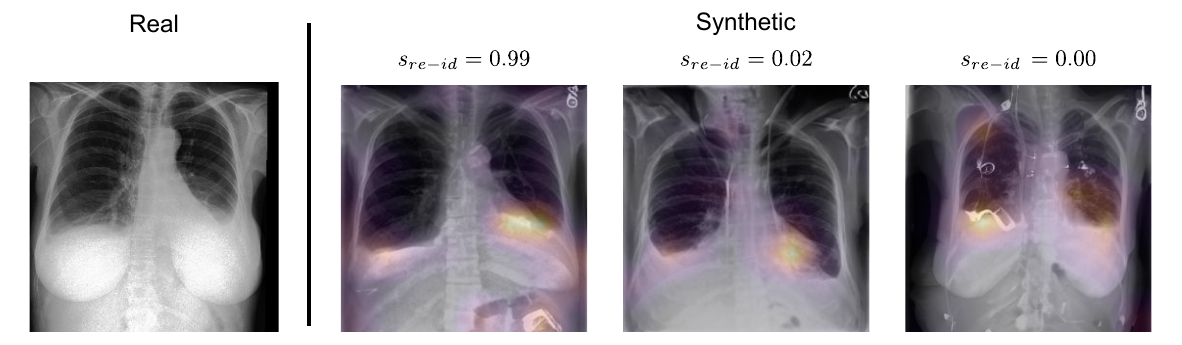}
    \caption{Grad-CAM++ heatmaps over the $\freid$. From left to right, the first synthetic image contains identifiable information about the real image in the areas towards the bottom of the lungs. The other two images do not contain identifiable information as defined by the re-identification score $s_{re-id}$.}
    \label{fig:grad_cam}
\end{subfigure}
\vfill
\begin{subfigure}[b]{\textwidth}
    \includegraphics[width=\textwidth]{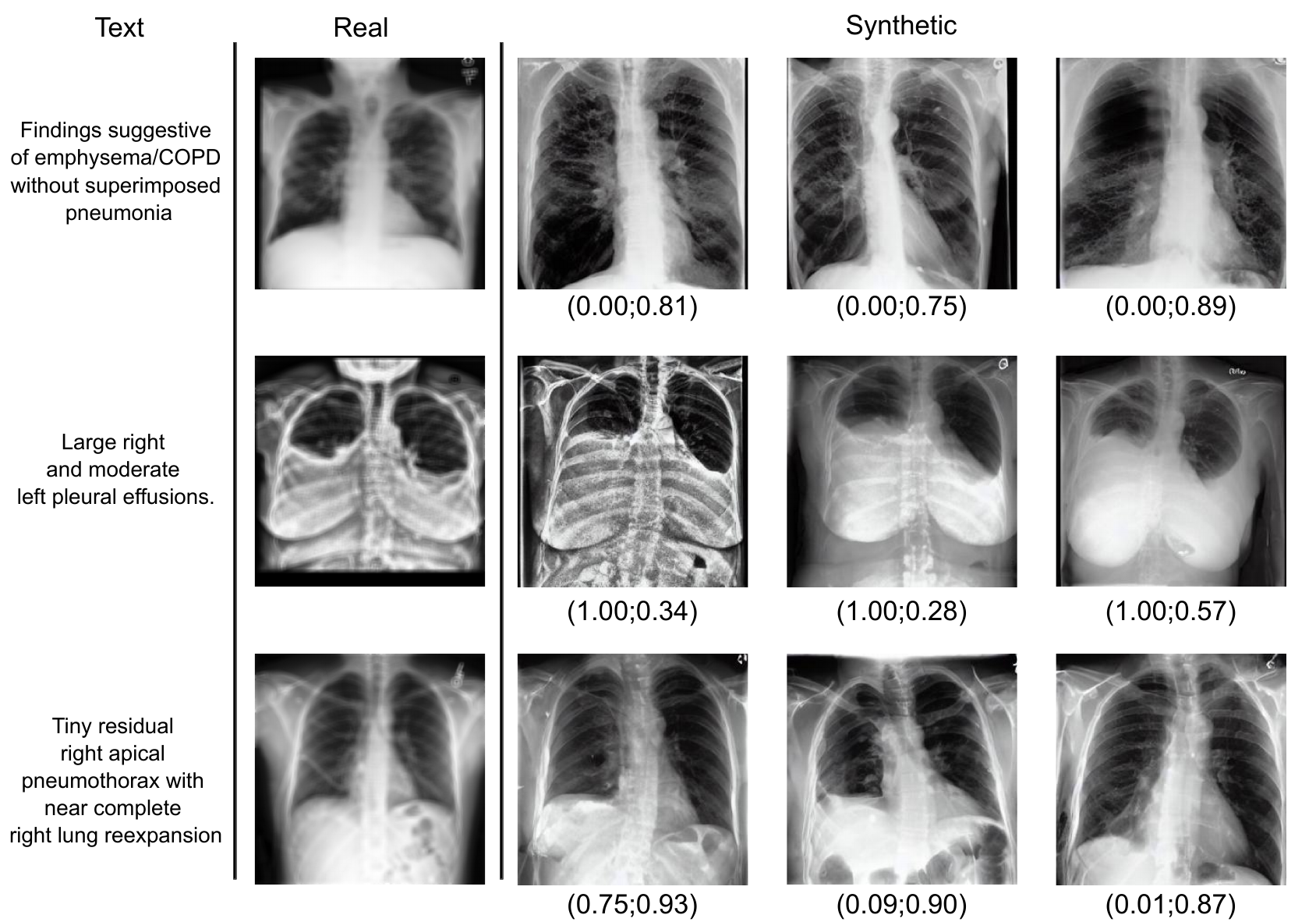}
    \caption{We display several synthetic images along with their prompt and real image associated with the prompt. We show below the synthetic images the re-identification score and text-to-image similarity in the format $(s_{re-id} ; \salign)$. In the first row, the synthetic images do not contain any identifiable information about the real image while corresponding fairly well to the text description. In the second row, all synthetic images contain identifiable information, despite having a different style/contrast. The bottom row displays synthetic images with good text alignment but some contain identifiable information and some do not.}
    \label{fig:visual_examples}
\end{subfigure}
\caption{Illustration showing how synthetic images, obtained by using the same prompt of the real image as conditioning, relates to the real image as well as the prompt. We try to mitigate privacy risks in these illustrations because the license of MIMIC-CXR \footnote{\url{https://physionet.org/content/mimic-cxr/view-license/2.0.0/}} does not allow sharing data. In \ref{fig:grad_cam}, the ``Real'' image is actually a synthetic image that is very similar to the original image. In \ref{fig:visual_examples}, we blurred the real images.}
\end{figure*}

\noindent\textbf{Visualising of How Synthetic Examples are Memorised.~}
We now focus on analysing Grad-CAM++ \cite{chattopadhay2018grad} heatmaps of the $\freid$ network in Figure \ref{fig:grad_cam} and some samples of synthetic images and their associated conditioning information in Figure \ref{fig:visual_examples}. For the explainability heatmap, we used the second-order gradients of the fourth layer of the ResNet-50 (which is a common choice for this architecture\footnote{\url{https://github.com/jacobgil/pytorch-grad-cam\#chosing-the-target-layer}}).

\subsection{Privacy Distillation}

We now show empirically that $\distillmodel$ indeed reduces re-identification risk. We also assess whether $\distillmodel$ is able to produce useful synthetic
datasets for downstream tasks. We train classifiers $\fclass$ (see section \ref{downstream}) on $\Dreal$, $\Dsyn$, capped at 40,000 images, and a dataset $\Ddistill$ of 40,000 images sampled from $\distillmodel$, our distilled model, using the training set prompts. We measure the AUC on the 3,101 images of our test set. The results are reported in table \ref{distillation_performances}, in addition to the re-identification ratio $R_{re-id}$ and the $\salign$ score. 

Training a model on synthetic data slightly affects performance (AUC and $\salign$ decrease), but the resulting value is still comparable to the literature \cite{Jacenkow2022}. Nonetheless, the re-identification ratio between the initial and the distilled models is decreased by more than 3-fold. We hypothesise that filtering re-identifiable data might also filter out unique phenotypes, more prone to be memorised \cite{Carlini2023ddpm}, resulting in reduced model generalisability and performance. 

\begin{table}[]
\caption{Privacy Distillation performance: memorisation ratio, predicted AUC for the classifier on the real test set and $\salign$ score.}
\begin{center}
\begin{tabular}{ lccc } 
 \toprule
Dataset & $R_{re-id}$ $\downarrow$ &  $\fclass$ AUC $\uparrow$  &  $\salign$ $\uparrow$ \\ 
 \midrule
 $\Dreal$ & - & $0.863$ & $0.698_{0.259}$ \\ 
 $\Dsyn$ & $4.24 \%$ & $0.830$ & $0.645_{0.219}$ \\ 
 $\Ddistill$ & $\mathbf{1.34} \%$ & $0.810$ & $0.611_{0.272}$ \\ 
 \bottomrule
\end{tabular}
\label{distillation_performances}
\end{center}
\end{table}

\section{Discussion and conclusion}

This study has demonstrated that the application of \textit{privacy distillation} can effectively reduce the risk of re-identification and leakage in latent diffusion models without excessively compromising the downstream task performance. In line with other approaches, such as differential privacy \cite{Dockhorn2022}, there is a trade-off between privacy and quality. In our proposed method, the trade-off is between the degree of filtering (more privacy) and downstream model utility (less privacy). Additionally, \textit{privacy distillation} can be applied iteratively by adding more filtering-sampling-training steps, an approach that should be the subject of future experiments.

This approach has the potential to facilitate the sharing of medical imaging generative models for fine-tuning and subsequent use. Other downstream tasks, such as segmentation, which we could not do due to the absence of ground truth masks, should also be contemplated to further explore the impact of the filtering. While in this paper, we use a text-to-image synthesis network and rely on a re-identification metric devised specifically for X-ray imaging, our approach could be applied to other imaging modalities and conditioning types, replacing the data, model and re-identification metric. 

Please note that while the proposed method requires training a second model on a filtered dataset, the filtering approach can be applied alone to reduce the re-identification risk of synthetic datasets (in the case where realising a synthetic dataset instead of a model is enough). This post hoc filtering procedure acts as a model auditor \cite{pmlr-v162-alaa22a} and can be used to improve any generative model. This is of particular interest when retraining a model is expensive or imposing certain properties is impractical.

\subsection{Limitations}

While \textit{privacy distillation} significantly reduces re-identifiability, it is important to note that some minor risks may still exist. We highlight that the ability to measure these risks is bounded by the accuracy and generalisation capabilities of the chosen measure of re-identification $s_{re-id}$.
Ensuring privacy preservation is a task that depends on various assumptions, such as the definition of ``re-identifiability'', which is limited by the choice of the metric and threshold. Further research should explore alternatives with a combination of metrics, such as the Kullback-Leibler distance or a loss-based score, as proposed by Hu et al.\ \cite{Hu_privacy_dms}.  

\clearpage

\nocite{langley00}

\bibliography{example_paper}
\bibliographystyle{icml2023}



\end{document}